\documentclass[times,twocolumn,final,authoryear]{elsarticle}

\usepackage{prletters}
\usepackage{framed,multirow}

\usepackage{amssymb}
\usepackage{latexsym}

\usepackage{url}
\usepackage{xcolor}
\usepackage{graphicx}
\usepackage{subfigure}
\usepackage{amsmath}
\usepackage{amssymb}
\usepackage{amsxtra}
\usepackage{braket}
\usepackage{array}
\usepackage{color}
\usepackage{url}
\usepackage{placeins}

\usepackage{algorithmic}
\usepackage{algorithm2e}
\usepackage{threeparttable}

\usepackage{times}
\usepackage{epsfig}
\usepackage{balance}
\usepackage{latexsym}
\usepackage{enumerate}
\usepackage{multirow}
\usepackage{pifont}
\usepackage{marvosym}
\usepackage{ifsym}

\usepackage{times}
\usepackage{braket}

\definecolor{newcolor}{rgb}{.8,.349,.1}

\journal{Pattern Recognition Letters}

\begin{document}

\thispagestyle{empty}

\begin{table*}[!th]

\begin{minipage}{.9\textwidth}
\baselineskip12pt
\ifpreprint
  \vspace*{1pc}
\else
  \vspace*{-6pc}
\fi

\noindent {\LARGE\itshape Pattern Recognition Letters}
\vskip6pt

\noindent {\Large\bfseries Authorship Confirmation}

\vskip1pc

{\bf Please save a copy of this file, complete and upload as the
``Confirmation of Authorship'' file.}

\vskip1pc

As corresponding author
I, \underline{Zhuo Xu},
hereby confirm on behalf of all authors that:

\vskip1pc

\begin{enumerate}
\itemsep=3pt
\item This manuscript, or a large part of it, \underline {has not been
published,  was not, and is not being submitted to} any other journal.

\item If \underline {presented} at or \underline {submitted} to or
\underline  {published }at a conference(s), the conference(s) is (are)
identified and  substantial \underline {justification for
re-publication} is presented  below. A \underline {copy of
conference paper(s) }is(are) uploaded with the  manuscript.

\item If the manuscript appears as a preprint anywhere on the web, e.g.
arXiv,  etc., it is identified below. The \underline {preprint should
include a  statement that the paper is under consideration at Pattern
Recognition  Letters}.

\item All text and graphics, except for those marked with sources, are
\underline  {original works} of the authors, and all necessary
permissions for  publication were secured prior to submission of the
manuscript.

\item All authors each made a significant contribution to the research
reported  and have \underline {read} and \underline {approved} the
submitted  manuscript.
\end{enumerate}

Signature\underline{:Zhuo Xu} Date\underline{:01/09/2022}
\vskip1pc

\rule{\textwidth}{2pt}
\vskip1pc

{\bf List any pre-prints:arXiv}
\vskip5pc

\rule{\textwidth}{2pt}
\vskip1pc

{\bf Relevant Conference publication(s) (submitted, accepted, or
published):NA}
\vskip5pc

{\bf Justification for re-publication:NA}

\end{minipage}
\end{table*}

\clearpage
\thispagestyle{empty}
\ifpreprint
  \vspace*{-1pc}
\fi

\begin{table*}[!th]
\ifpreprint\else\vspace*{-5pc}\fi

\section*{Research Highlights (Required)}

To create your highlights, please type the highlights against each
\verb+\item+ command.

\vskip1pc

\fboxsep=6pt
\fbox{
\begin{minipage}{.95\textwidth}
It should be short collection of bullet points that convey the core
findings of the article. It should  include 3 to 5 bullet points
(maximum 85 characters, including spaces, per bullet point.)
\vskip1pc
\begin{itemize}

 \item{We formalize the writing style in documents as the Role-Rank Distribution between the argument roles, positions, and the event triggers.}

 \item{We propose a new document-level event extraction model based on our defined Role-Rank Distribution.}
 
 \item{We capture the Role-Rank Distribution by training our model with the event extraction task.}
 
 \item{We verifies the captured Role-Rank Distribution contains valuable information that could improve the performance of the event extraction task.}

\end{itemize}
\vskip1pc
\end{minipage}
}

\end{table*}

\clearpage

\ifpreprint
  \setcounter{page}{1}
\else
  \setcounter{page}{1}
\fi

\begin{frontmatter}

\title{Writing Style Aware Document-level Event Extraction}
\author[CUFE]{Zhuo \snm{Xu}}
\address[CUFE]{Department of Computer Science, Central University of Finance and Economics, Beijing, 102206, China\\}

\author[CUFE]{Yue \snm{Wang}\corref{mycorrespondingauthor}}
\cortext[mycorrespondingauthor]{Yue Wang is the corresponding author.}
\ead[url]{wangyuecs@cufe.edu.cn}

\author[CUFE]{Lu \snm{Bai} }
\author[CUFE]{Lixin \snm{Cui} }

\begin{abstract}
Event extraction, the technology that aims to automatically get the structural information from documents, has attracted more and more attention in many fields. Most existing works discuss this issue with the token-level multi-label classification framework by distinguishing the tokens as different roles while ignoring the writing styles of documents. The writing style is a special way of content organizing for documents and it is relative fixed in documents with a special field (e.g. financial, medical documents, etc.). We argue that the writing style contains important clues for judging the roles for tokens and the ignorance of such patterns might lead to the performance degradation for the existing works. To this end, we model the writing style in documents as a distribution of argument roles, i.e., Role-Rank Distribution, and propose an event extraction model with the Role-Rank Distribution based Supervision Mechanism to capture this pattern through the supervised training process of an event extraction task. We compare our model with state-of-the-art methods on several real-world datasets. The empirical results show that our approach outperforms other alternatives with the captured patterns. This verifies the writing style contains valuable information that could improve the performance of the event extraction task.
\end{abstract}
\begin{keyword}
\KWD Document-level event extraction\sep Writing style\sep Role-Rank Distribution.
\end{keyword}
\end{frontmatter}

\section{Introduction}
Event extraction (EE)~\citep{1} is an important mission to get event factors~\citep{22}, including triggers~\citep{17} and arguments~\citep{18}, from real-world corpora. Usually, the event triggers~\citep{17} are mentions that express the causes and types for events and the event arguments are named entities (e.g. person names, company names, and locations)~\citep{2} that play critical roles in an event. As a method to analyze texts, the EE system is wildly used in applications like Information Extraction~\citep{3}, Question Answer~\citep{4}, etc. The mainstream works~\citep{8,9} divide into sentence-level and document-level event extraction according to the different scopes of the inputs. Further, since it is a combinatorial optimization problem to align entity labels with tokens, many existing works~\citep{9,6,22} try to include several potential features to find a good mapping for tokens and their labels. Literature~\citep{9,6} leverage semantic features in their models. And some methods~\citep{22} adopt pre-defined information or pre-trained data to improve the performance of their models. The aforementioned works could be considered as the method to reform the token-level multi-label classifiers which distinguish the tokens in sentences as different roles.


Recently, researchers and practitioners apply event extraction technologies to aid the studies in many fields~\citep{6,35,36}. For instance, ~\cite{37} apply the event extraction method to enhance finance-related research performed by financial analysts. ~\cite{36} apply the event extraction results from medical reports to make related data easily accessible and improve medical decisions. These studies show that high-quality event extraction results have the potential to boost studies in special fields. However, since the contents of the documents in special fields  (e.g. finance, medical, etc.) organize relatively fixed, the quality of the event extraction results might be degraded by applying a general model. This has a negative influence on the development of consequential studies.

\begin{figure}[ht]
\centering
\subfigure[The probability distribution of event roles in the equity pledge documents.]{
\begin{minipage}{.5\textwidth}
\label{fig:motivation:a}
\includegraphics[width=0.77\textwidth]{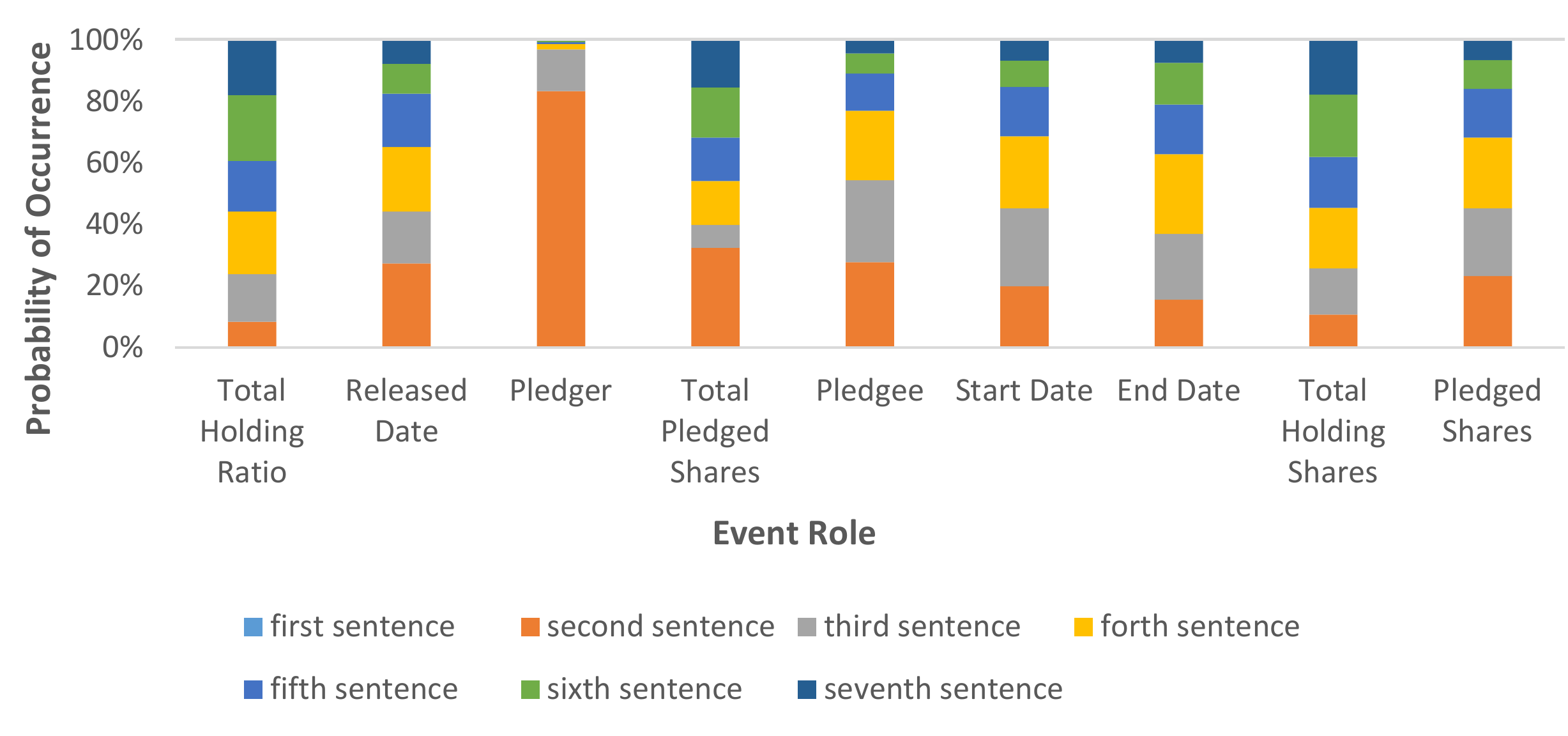}
\centering
\end{minipage}
}
\subfigure[The probability distribution of event roles in the equity repurchase documents.]{
\label{fig:motivation:b}
\begin{minipage}{.5\textwidth}
\includegraphics[width=0.77\textwidth]{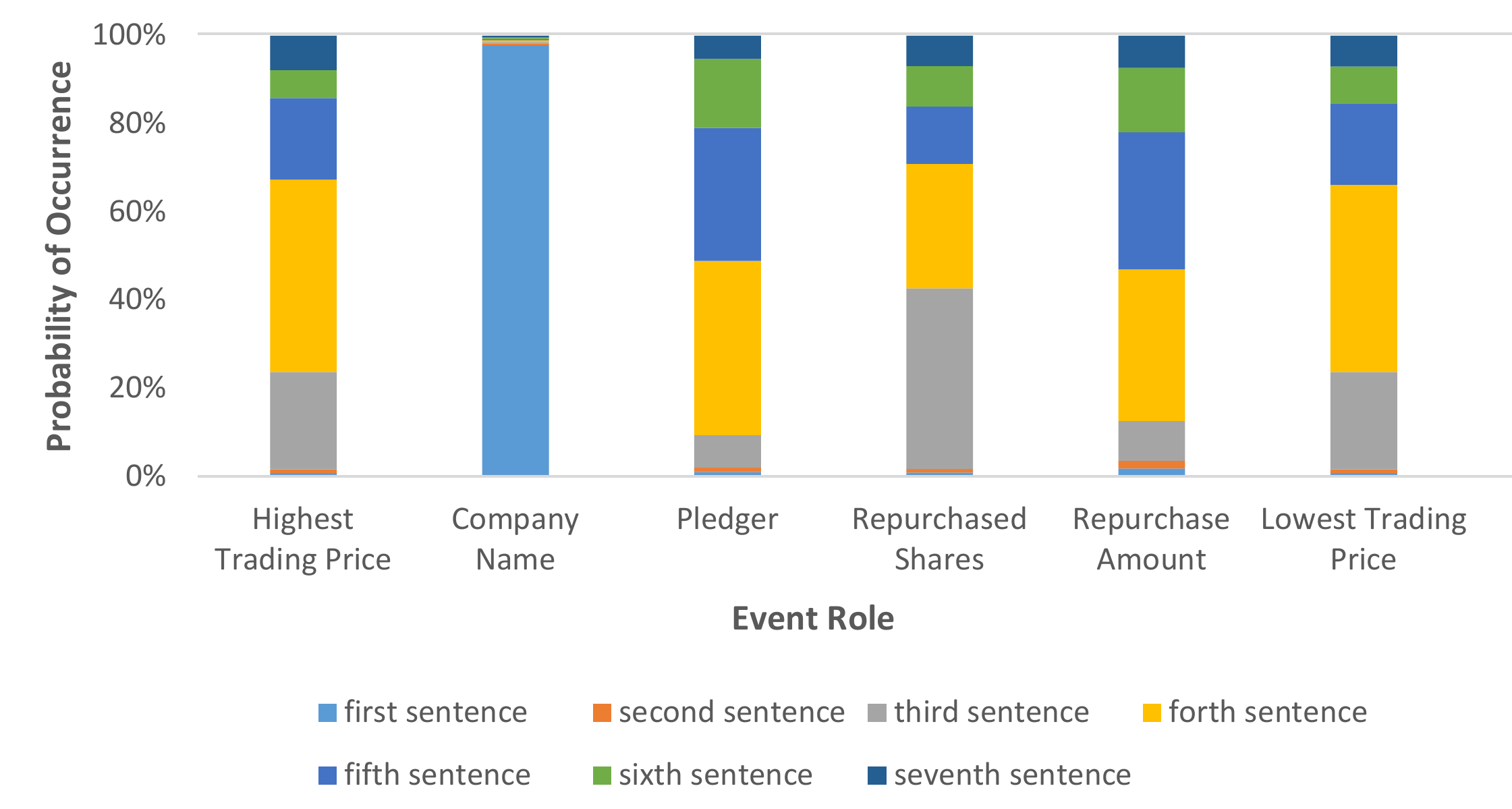}
\centering
\end{minipage}
}
\caption{We define the writing style as a distribution of event roles in documents (will be depicted in Section \ref{sec:RRD}). This figure shows two distributions of financial documents in different event roles (pledge and repurchase). We observe that these kinds of distributions are relatively fixed to the documents in documents of specific event types.} \label{fig:motivation}
\end{figure}

The writing style~\citep{11} describes a special way to organize the contents for a document in a specific field. It is widely used in different fields such as recognizing stylistic deception~\citep{23}, predicting the success of literary works~\citep{24}, etc. In order to incorporate the writing style in the event extraction task, we formalize it as, Role-Rank Distribution (RRD) distribution, a distribution that describes the roles and positions of the named entities in a document. In our experiment (cf. Fig.~\ref{fig:motivation}), we observe that the RRD distributions retain relatively fixed for documents of specific event types in the financial corpus CFA (Chinese Financial Announcement ~\citep{10}). This indicates that the writing style have a strong connection with the event factors in the financial documents with given event types and it might has the potential to help the performance of the event extraction task. However, as far as we know, none of the existing works consider this special pattern or how to leverage it in their models.



To this end, we leverage the RRD distribution in a proposed joint event extraction model. Since our model considers the relationships among the roles, positions for arguments and the corresponding event triggers in the given documents, it improves the performance of extracting the named entities and the event triggers simultaneously.

We test our model on real-world datasets and our model excels the state-of-the-art event extraction works with the same training set. This indicates that our method captures useful writing styles in helping the event extraction task in specific fields.

In summary, the main contributions of our work are follows:
\begin{itemize}
    \item To our knowledge, we are the first to discuss the relationship between the argument roles, positions and the event triggers in the field-oriented documents (cf. financial documents) and we formalize this relationship as the Role-Rank Distribution (RRD) distribution in the event extraction task.
    \item We propose a model, the Event Extraction via Field-Role-Rank Distribution (EEFRRD), which extracts event information from documents by leveraging the Role-Rank Distribution (RRD) distribution towards a specific field.
    \item Our experiments on the financial document datasets show that, by integrating the RRD pattern, our method achieves better performance on the Document-level Event Extraction (DEE) tasks than the other related alternative methods. We also observe that this performance improvement is further significant when dealing with financial documents that have strong RRD patterns.
\end{itemize}

The remainder of paper is as follows: Section 2 introduces some definitions about event extraction and argument distribution. Section 3 proposes our model and gives the details of the RRD distribution. Section 4 conducts sufficient comparison experiments of our method and other alternatives on several real-world financial datasets. Section 5 concludes this paper.

\section{Preliminaries}

We give the formal definition for the details of the event extraction tasks and the definition of our problem in this section.

\subsection{Event Extraction}

To start with, we provide several key notations in Table~\ref{table:notations} by following~\citep{9,10}.

\begin{table}[ht]
\caption{The notations related in this work.}
\centering
\begin{tabular}{p{.1\textwidth}|p{.33\textwidth}}
\hline
Notation&Description\\
\hline
$a$&The event argument, it is the named entities that plays a necessary role in an event.\\
\hline
$r$&The event role, it is a field for a subset of event arguments.\\
\hline
$e$&The event type, it is a generalization of some similar events which are different from others.\\
\hline
$c=\langle{r},a\rangle$&The event record, it is a couple that is composed of a event role and one of its corresponding arguments. \\
\hline
$T=\langle{e}, C\rangle$&The event table, it is also a couple consists of the type of an event $e$ and a set of event records $C$.\\
\hline
\end{tabular}
\label{table:notations}
\end{table}

\begin{figure}
\centering
\includegraphics[width=0.5\textwidth]{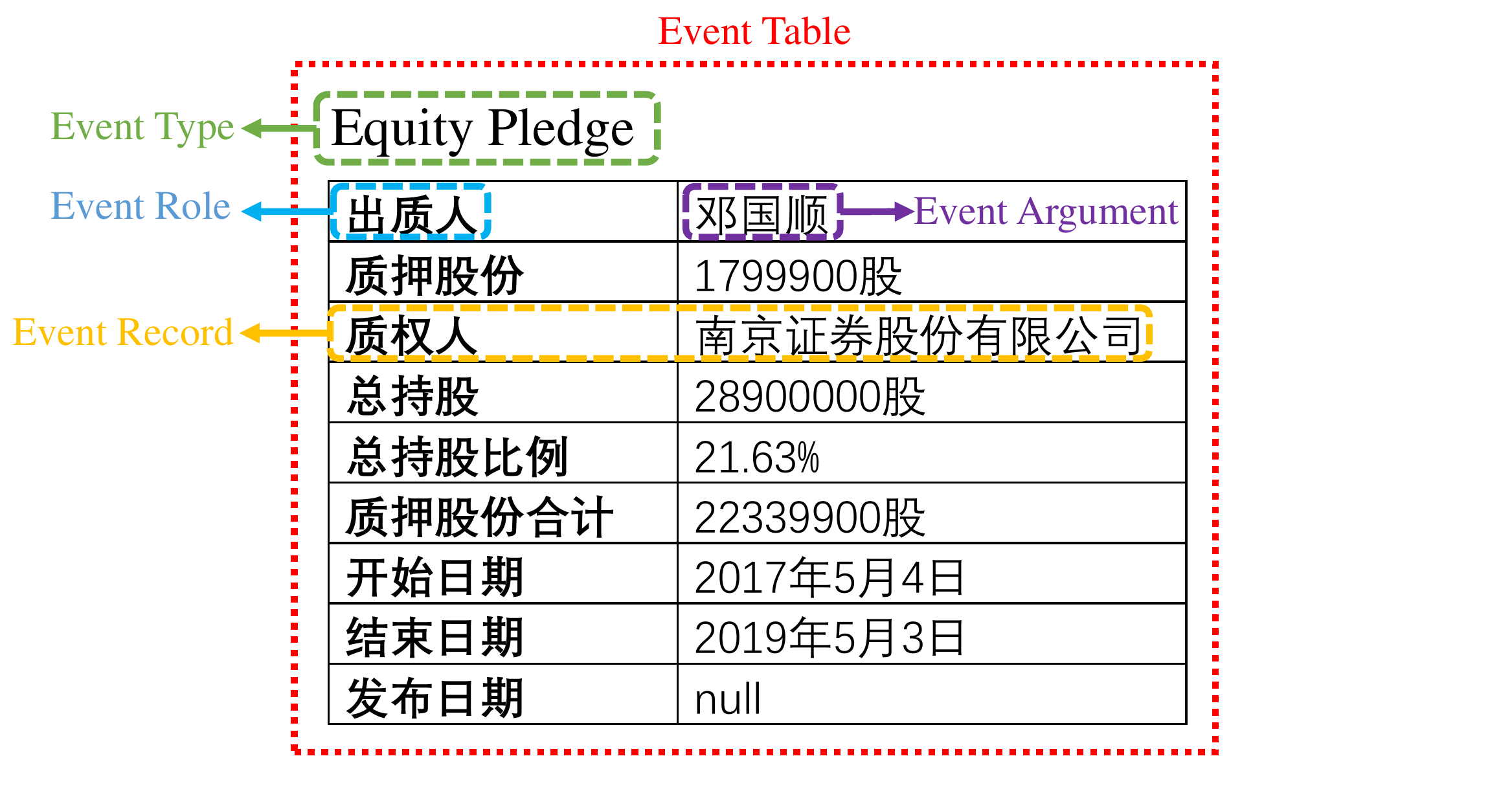}
\caption{An example of the extracted event table of the event extraction task.} \label{fig_definition}
\end{figure}

To illustrate the aforementioned conceptions, we show a real-world financial event table in Figure~\ref{fig_definition}. Generally, the event extraction is a task to fill the right blanks of an event table with mentions in a document. This event table contains structural information such as  event types, event roles, and event arguments.

Recent event extraction works divide into the sentence-level and the document-level methods that differ by different scopes of the input contents.

\subsubsection{Sentence-level Event Extraction (SEE).} The Sentence-level Event Extraction models aim at extracting the named entities~\citep{16} from sentences. Sequence-to-sequence (Seq2Seq)~\citep{15,29,30} model is a mainstream method to implement the SEE task. The idea~\citep{15} of Seq2Seq SEE models is to translate a sentence (or the token sequence) to a tag sequence in the BIO schema. Given a set of sentences $D$ and a set of pre-annotated label sequence $Y$; suppose for an arbitrary sentence $s=\{w_1,w_2,...,w_n\}$ ($\forall{s}\in{D}$, $w_i$s are tokens), there is a corresponding label sequence $g(s)=\{y_1,y_2,...,y_n\}$ ($|s|=|g(s)|$, $g(s)\in{Y}$); Then the task of Seq2Seq SEE is to find an optimal tagger function $\phi(s|\theta)$ that minimizes the following loss function:
\begin{equation}
\label{loss_1}
\mathcal{L}_1 =\sum_{\forall i \in [1,N_w]} \sum_{\forall y\in Y}-p_i log(\hat{p_i})
\end{equation}

where $\hat{p_i} = \hat{Pr}(y_i|w_i)$ is the probability to annotate a token $w_i$ as the tag $y_i$ by $\phi$, and $p_i = Pr(y_i|w_i)$ is the probability of an oracle model to annotate the same token as the tag $y_i$.


As an auto-annotation method, SEE models~\citep{29} extract event arguments within the scope of a sentence, and some~\citep{32} reach excellent performance. However, in a real-world document, the factors (arguments, roles, event type, etc.) of an event may scatter in the whole document rather than within a single sentence. This raises the argument-scattering issue~\citep{9} of how to get the optimal scattered factors of an event for the current SEE methods.

\subsubsection{Document-level Event Extraction.}

Document-level Event Extraction (DEE)~\citep{10} extends the scopes of the input contexts from sentences to documents. This gets more candidate arguments with roles and thus could generate a complete event table (cf. Figure~\ref{fig_definition}) from a document.
Recent works~\citep{9,10,31} have explored to build models on the document level.  \cite{9} explore to construct DEE framework on two stages: tagging sequence by SEE and utilizing multi-sentence to pad missing information.
 \cite{10} try to add generate an entity-based directed acyclic graph to fulfill DEE process.
Some efforts try to add contextual features, such as syntactic features~\citep{21} to help identify event types from the texts.

\subsection{Problem Formulation}

Our task is to extract the complete structural information (arguments, roles, event type, etc.) of an event from a document. We formalize this document-level joint event extraction process as follows.

Given a document $d$ ($d=\{s_1,s_2,s_3,...\}$), where $s_i$s are sentences. Then our target is to find an optimal event table $T'$ with the following method.
\begin{equation}
    T'= \langle\mathcal{E}(d),\bigcup_{s_i\in{d}}{f(s_i)}\rangle,
\end{equation}
where $\mathcal{E}$ is a function to identify the type of event based on the input of $d$.

As is discussed in the Introduction, we observe that a significant difference in the writing styles for documents of various event types. However, this pattern is ignored by most of the existing works. We try to leverage this principle to improve the performance of our task in the consequential sections.

\section{Our Proposed Model}

\subsection{Overview of Our Model}

\begin{figure}[ht]
\centering
\includegraphics[width=.5\textwidth]{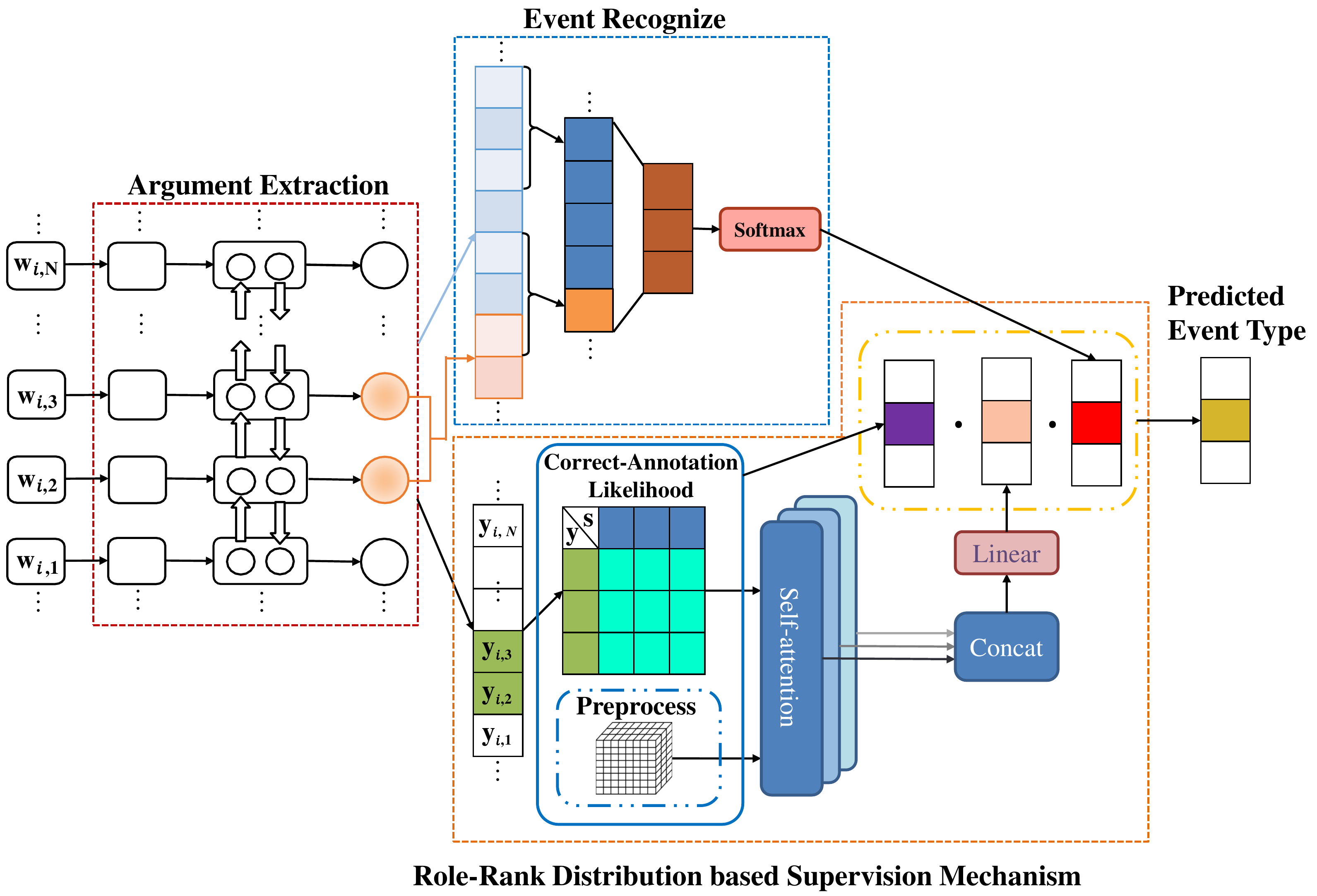}
\caption{The training framework of  DEE model with RRDSM } \label{fig2}
\end{figure}
Figure~\ref{fig2} shows the main framework of our model. First, it extracts all the arguments with a sentence-level labeling module. Then, it utilizes a neural network classifier to distinguish the event type.  After the aforementioned two processes, our system obtains the candidate event records (a set of event records). We generate the target event table by filtering the obtained event records based on the distribution of roles to the arguments in documents of different event types. To this end, we formalize the writing style as the Role-Rank Distribution and propose the Role-Rank Distribution based Supervision Mechanism (RRDSM) to measure the likelihood of a document belonging to a specific event type. This mechanism further improves the performance of our model.

As is illustrated in Figure~\ref{fig2}, our argument extraction module applies a BiLSTM-CRF~\citep{34} based model to realize argument extraction from sentence sequences. Then, our event recognition module uses a CNN~\citep{21}-based model to classify the event types for documents. In the last, we propose a self-attention augmented module by leveraging the proposed Role-Rank Distribution to revise the final event type prediction. We elaborate on the details of all these modules in the following sections.


\subsection{Role-Rank Distribution}
\label{sec:RRD}


We formalize the writing styles as a statistical distributions about the argument roles, argument positions in sentences and event triggers. To facilitate the discussion, we provide the related notations in the following.

\subsubsection{Role-Rank Score.}
Given a set of documents $D$, a set of argument roles $R$. Then the Role-Rank Score $p(r|i)$ $\forall{r}\in{R},\forall{i}\in{[1,N]}$, $N$ is the maximum length of the sentences in $D$) is a conditional probability for a specific event role $r$ that appears in the $i$-th sentences of all the documents in $D$.

In order to capture the Role-Rank Scores for documents with different event types, we extend the it to the following form.

\subsubsection{Role-Rank Distribution.}
Given a set of events $E$, a set of documents $D$, a set of argument roles $R$. The Role-Rank Distribution is an $\mathbb{R}^{|E|\times{N}\times|R|}$ tensor $P$, where each element of it $p(r|i,e)$ is a conditional probability of an argument role $r$ ($\forall{r}\in{R}$) appearing in the $i$-th sentence belonging to the documents of event type $e$ ($\forall{e}\in{E}$).

The Role-Rank Distribution reveals the ternary relationship of the event roles and the order of sentences in documents under different event types. From the observation in the Introduction, this distribution could be rather different with various events. Therefore, it might be potentially useful to predict the event roles by giving a sentence order and an event type. In a real-world corpus, a complete Role-Rank Distribution for a set of document $D$ under all the event types could be extremely huge ($|E|\times{N}\times|R|$). This might degrade the efficiency of our prototype system. To this end, we preprocess and storage an Role-Rank Distribution as the $\mathbb{R}^{|E|\times{N}\times|R|}$ tensor $P$. In the following sections, we also use the $\mathbb{R}^{{N}\times|R|}$ slice matrix $P_e$ refers to a slice of the Role-Rank Distribution towards a specific event type $e$ ($\forall{e}\in{E}$).

\subsubsection{Preprocessing the Role-Rank Distributions.}
In order to reduce the time complexity for our prototype system, we obtain these Role-Rank Distributions by a preprocessing process before the training of our model. With the preprocessed results, the Role-Rank Distributions of the training data could be accessed whenever required by the training process of our model. We describe this process in Algorithm~\ref{algorithm:datacube}.

\begin{algorithm}
    \label{algorithm:datacube}
  \caption{Preprocessing the Role-Rank Distribution}
  \KwData{a document set $D$s, their event table set $T$s}
  \KwResult{The Role-Rank Distribution tensor $P=\{p(r|i,e)\}_{|E|\times{N}\times{|R|}}$}
  Initialize the tensor $P=\{p(r|i,e)\}_{|E|\times{N}\times{|R|}}$ with all ones.\\
    \For{$\forall{e}$ in $E$}{
        \For{$\forall{d}$ in $D$}{
        \For{$\forall{r}$ in $R$}{
         i$\leftarrow$find the order number of the sentence including $a$ in $d$\\
         $p(r|i,e)\leftarrow{p(r|i,e)+1}$\\

        }
        }
        \For{$\forall{r}$ in $R$}{
        $t_{r}\leftarrow{\sum_{j=0}^{N-1}p(r|j,e)}$\\
        \lFor{$i$ in $[0,N-1]$}{
         $p(r|i,e)\leftarrow{\frac{p(r|i,e)}{t_{r}}}$
        }
        }
    }
    \textbf{Output: }$P=\{p(r|i,e)\}_{|E|\times{N}\times{|R|}}$
\end{algorithm}
In our experiment, the preprocessing of Algorithm~\ref{algorithm:datacube} usually saves 12.8\% of the training time of our prototype system by comparing to the original on-line processing.

\subsection{Role-Rank Distribution Based Supervision Mechanism}

To incorporate the preprocessed Role-Rank Distribution to supervise our model, we propose the Correct-Annotation Likelihood to evaluate the plausibility of a document to a specific event type.
The core idea of this method is to analyze the possibility of the output tag sequences from a Seq2Seq SEE module by referring the preprocessed Role-Rank Distribution of the training set. This requires a Role-Rank Distribution of the Seq2Seq SEE outputs. However, since the Seq2Seq SEE only outputs tag sequences, there is a gap between the result of the Seq2Seq SEE module and the preprocessed Role-Rank Distribution of the training set. To this end, we propose the Tag-to-Role Transition generate the required Role-Rank Distribution from the Seq2Seq SEE results.

\subsubsection{Tag-to-Role Transition}
Given a set of documents $D$, a set of tags $Y$, and a set of event roles $R$; suppose $s$ is a sentence in document $d$ ($\forall{d}\in{D}$), $g(s)$ is the corresponding tag sequence for $s$. Then the Tag-to-Role Transition $W$ is an $\mathbb{R}^{|Y|\times|R|}$ matrix, where each of its element $w_{t,r}$ is the frequency that a tag $t$ belongs to role $r$ in $D$.

With a set of tag sequences outputted from the Seq2Seq SEE module, the Tag-to-Role Transition helps to transform the distribution of the tags to the sentence order to a role-rank distribution. Formally, a role-rank distribution $P'$ can be obtained through the transformation in Equation~\ref{eq:transform}.
\begin{equation}
\label{eq:transform}
   P'=P_y(d)\cdot{W},
\end{equation}
where $P_y(d)$ is an $\mathbb{R}^{N\times{|Y|}}$ matrix that refers to the distribution between the tags and the sentence orders in document $d$. With the obtained Role-Rank Distribution from the annotated result of Seq2Seq SEE module, we define the Correct-Annotation Likelihood as the following.

\subsubsection{Correct-Annotation Likelihood.} Let $P$ be a tensor of preprocessed Role-Rank Distribution and $P'$ be the Role-Rank Distribution counted from the annotated result of the SEE module. Suppose $P_e$ is the slice matrices of $P$ which is divided by event type. Then the Correct-Annotation Likelihood is computed as the following Equation.
\begin{equation}
\label{eq:likelihood}
   l_e = similarity(P',P_e) = \frac{P'\cdot P_e}{||P'||\times||P_e||}  \forall{e}\in{E},
\end{equation}
where $e$ is a specific event type $\forall{e}\in{E}$, $||\cdot{||}$ is the $L^2$-norm. Note that, although we use the Cosine similarity in this work, this metric could also be  other similarity method.

Our model uses the obtained Correct-Annotation Likelihoods as the weights to revise the event prediction. The experiment shows that this method is effective in improving the performance of the event prediction. However, we further observe deeper potential principles that might lie behind the Role-Rank Distribution in our experiment. Concretely, as is illustrated in Figure~\ref{fig:motivation}, we observe that although most of the sentence distributions of different argument roles are significantly different, some of them might rather alike (c.f. the distributions of role ``Pledger'' and ``Repurchase Amount'' in Figure~\ref{fig:motivation:b}). To distinguish the roles of the arguments with similar distributions, we further explore the Role-Rank Distribution in different event types and found that their distribution may be significantly different under various event types (c.f. the role ``Pledger'' in event types ``equity pledge'' and ``equity repurchase '' in Figure~\ref{fig:motivation:a}). In the next section, we propose a self-attention method to leverage this latent principle from the observation to further improve the performance of our model.

\subsubsection{Self-attention Augmented Event Identification.}

 We propose a self-attention based module to further revise the event type prediction results by leveraging the latent difference among the same event types.

 Given a  document $d (\forall{d}\in D)$ , a set of event types $E$ and a set of event roles $R$. Then, the event attention $a_e$ for event $e$ ($\forall{e}\in{E}$) is computed as the following Equation.
 \begin{equation}
     \label{eq:attention}
     a_e= Softmax(\frac{P'P^T_e}{\sqrt{|R|}})Q,
 \end{equation}
where $Q$ is a distribution of event roles to event type and it can be computed as Equation~\ref{eq:trans_r_to_e}.
 \begin{equation}
     \label{eq:trans_r_to_e}
     Q = P'W',
 \end{equation}
 where $W'=\{w_{r,e}\}_{|R|\times|E|}$ is the transition matrix from event roles to event types. Its element $w_{r,e}$ is the frequency that a role $r$ belongs to a event type $e$ in $D$; $P'$ is the Role-Rank Distribution counted from the annotated result of the SEE module; $P_e$ is the slice matrices of the preprocessed $P$.


In order to improve the efficiency in computing the attentions, we concatenate the attention results to form the following tensor.
\begin{equation}
    \label{eq:weight}
    A = Linear(a_{e_1}\oplus{a_{e_2}}\oplus...\oplus{a_{_{e_n}}})
\end{equation}
where $e_i$s are various events ($\forall{e_i}\in{E}$) and the function $Linear$ is a linear transformation or full connected layer to transform the output as the $\mathbb{R}^{|E|}$ column tensor.

\subsection{Complete task and Optimization}

Our complete task consists of two sub-tasks, i.e., the Sentence-level Event Extraction (SEE) and the Event Type Identification. The loss function $\mathcal{L}_1$ for the SEE task is defined in Equation \ref{loss_1}. The loss function $\mathcal{L}_2$ for the Event Type Identification is formalize as the following.


\begin{equation}
    \label{eq:loss2}
    \mathcal{L}_2 = \sum_{\forall{d}\in{D}}\sum_{\forall e \in E}- {p_{e,d}}\log((Softmax([l_{1},l_{2},...,l_{|E|}]\cdot A\cdot V))_{e}),
\end{equation}
where $p_{e,d}$ is the probability of an oracle model to classify document $d$ into the event $e$, $[l_{1},l_{2},...,l_{|E|}]^{T}$ is an $\mathbb{R}^{|E|}$ vector which is consisted of the Correct-Annotation Likelihoods of document $d$ toward various event types, $A$ is the attention results computed from Equation~\ref{eq:weight} and $V$ represents an $\mathbb{R}^{|E|}$ event extraction results from document $d$; Since the result of $Softmax(\cdot)$ function in Equation~\ref{eq:loss2} is an $\mathbb{R}^{|E|}$ vector, $Softmax(\cdot)_e$ is the $e$-th  element of the $Softmax$ result and it represents the probability of our model to classify document $d$ into the event $e$.

By combining the two sub-tasks, our final task is computed as following Equation.
\begin{equation}
    \label{loss}
    \mathcal{L} = (1-\lambda){\mathcal{L}_1} +\lambda{\mathcal{L}_2},
\end{equation}
where $\lambda$ is a parameter to adjust the weight of the event identification task. After optimizing this loss function, our system obtains the tagging results and the event type simultaneously. Therefore, after the training process, our system outputs an approximate optimal event table for the target documents.

\section{Experiment and Analysis}

\subsection{Dataset}
We use the CFA (Chinese Financial Announcement, 2008-2018) compiled by ~\cite{10} through all the experiments in this work. It is based on the knowledge base of remote supervision and expert summary to mark the text data. The event can be divided into three types in Table~\ref{dataset}: Equity Repurchase (ER), Equity Underweight (EU), and Equity Pledge (EP). These text data include major news events that have been disclosed and may have a huge impact on the behavior of companies and investors. The partition information for the dataset is shown below.

\begin{table}
\caption{A Summary for Our Dataset.}
\setlength{\tabcolsep}{4mm}{
\begin{tabular}{ccccc}
\hline
Event & Train & Dev  & Test & Total \\ \hline
ER    & 1,862  & 297  & 282  & 3,677  \\
EU    & 5,268  & 677  & 346  & 5,847  \\
EP    & 12,857 & 1,491 & 1,254 & 15,602 \\ \hline
All   & 19,987 & 2,465 & 1,882 & 25,126 \\ \hline
\end{tabular}}
\label{dataset}
\end{table}

This dataset is divided into training set, validating set and testing set in the proportion 8:1:1, which are used to train, validate and test the model.

\subsection{Comparison Baselines}
We compare our model with some baselines under the  DEE framework. The related models which adopted this framework are as follows.
\begin{itemize}
    \item JEE~\citep{31} is a model that depends among variables of events, entities, and their relations, and performs joint inference of these variables across a document.
    \item DCFEE-O is a model that gets one event record from a sentence based on DCFEE~\citep{9}.
    \item DCFEE-M is an improved model based on DCFEE, which can get combined event records from multi-sentences.
    \item Doc2EDAG~\citep{10} added an entity-based directed acyclic graph to optimize the structure of  DEE.
    \item EEFRRD-CAL is our completed model with the distribution supervision mechanism which only makes use of Correct-Annotation Likelihood.
    \item EEFRRD-SAEI is our completed model with RRDSM which adopts Self-attention Augmented Event Identification.
\end{itemize}

\subsection{Implementation Details}
In the tagging sequence, we adopt the Beginning-Inside-Outside (BIO) annotation schema for the candidate argument set extraction. To compare all methods fairly, we adopt the same BiLSTM module which has 4 hidden layers and 768 hidden dimensions. Moreover, all models are trained through the Adam optimizer. The learning rate is $1e^{-4}$. During the training, we set the batch size to 32 and the dropout to 0.5. In our model, we set 2 CNN layers where the size of the convolution kernel is 3. The style of pooling is max-pooling to mine the most significant feature. In calculate our loss function, we set $\lambda = 0.60$.

To evaluate the performance of our model, we adopt several prevalent metrics, which have been used to evaluate event extraction by comparing the predicted event table. From the predicted event table and the ground true table which belongs to the same event type from one document, we need to compare event records without replacement one by one and calculate true positive, false positive, and false negative ($TP$, $FP$ and $FN$ for short) statistics until there is no record left. At last, we can calculate precision, recall, and F1 scores ($P$, $R$, $F1$ for short). The equations are as follows:
$$P = \frac{TP}{TP + FP}$$
$$R = \frac{TP}{TP + FN}$$
$$F1 = \frac{2\cdot P \cdot R}{P + R}$$
\subsection{Experimental Results}
The results of the comparison experiment on the dataset are shown in Table~\ref{table_result}. We observe that with the RRDSM mechanism, our model EEFRRD-SAEI gets a better $F1$-score than the other  DEE models. This verifies that our proposed model indeed improves the performances of document-level event extraction.

\begin{table}
\caption{Comparison of different models on $P( \% )$,$R( \% )$ and $F1( \% )$.}
\setlength{\tabcolsep}{0.1mm}{
\resizebox{.475\textwidth}{!}{
\begin{tabular}{lcccccccccccc}
\hline
                &               & ER            &               &               & EU            &               &               & EP            &               &               & Avg.          &               \\
                & P             & R             & F1            & P             & R             & F1            & P             & R             & F1            & P             & R             & F1            \\ \hline
JEE         & 78.6          & 69.4          & 73.7          & 61.7          & 40.5          & 48.9          & 62.3          & 35.7          & 45.4          & 67.5          & 48.5          & 56.0          \\
DCFEE-O         & 84.5          & 81.8          & 83.1          & 62.7          & 35.4          & 45.3          & 64.3          & 63.6          & 63.9          & 70.5          & 60.3          & 64.1          \\
DCFEE-M         & 83.7          & 78.0          & 80.8          & 49.5          & 39.9          & 44.2          & 59.8          & 66.4          & 62.9          & 64.3          & 61.4          & 62.6          \\
Doc2EDAG        & 91.3          & 83.6          & 87.3          & 80.2          & 65.0 & 71.8          & 80.0          & 74.8 & 77.3          & 83.8          & 74.5 & 78.8          \\
EEFRRD-CAL & 94.1 & 84.0 & 88.7 & 85.4 & 63.6          & 72.9 & 82.7 & 74.5          & 78.4 & 87.4 & 74.0          & 80.0 \\
\textbf{EEFRRD-SAEI} & \textbf{94.7} & \textbf{86.4} & \textbf{90.4} & \textbf{87.3} & \textbf{67.2}          & \textbf{75.9} & \textbf{84.3} & \textbf{76.0}          & \textbf{79.9} & \textbf{88.8} & \textbf{76.5}          & \textbf{82.1} \\
\hline
\end{tabular}}}
\label{table_result}
\end{table}

In the experiment, it is hard for the model to recognize event types in multi-event documents. Table~\ref{table_event_classification} shows the event classification performance of different models.  EEFRRD-SAEI utilizes Role-Rank Distributions to obtain better extraction. As a necessary step of event table generation, this part has an obvious improvement under the supervision of the proposed RRDSM.

\begin{table}[]
\centering
\caption{Comparison of different models to classify event type on $F1( \% )$ on multi-event sets. }
\begin{tabular}{lcccc}
\hline
         & ER   & EU   & EP   & Avg. \\ \hline
JEE  & 48.5 & 42.4 & 63.6 & 52.4 \\
DCFEE-M  & 53.4 & 39.6 & 60.6 & 51.2 \\
Doc2EDAG & 68.4 & 64.6 & 72.5 & 68.5 \\
EEFRRD-CAL   & 71.8 & 63.2 & \textbf{78.1} & 71.0 \\
EEFRRD-SAEI   & \textbf{73.0} & \textbf{66.3} & 76.9 & \textbf{72.1}
\\ \hline
\end{tabular}
\label{table_event_classification}
\end{table}

\subsection{Sensitivity Study}
\begin{figure}
\centering\includegraphics[width=0.5\textwidth]{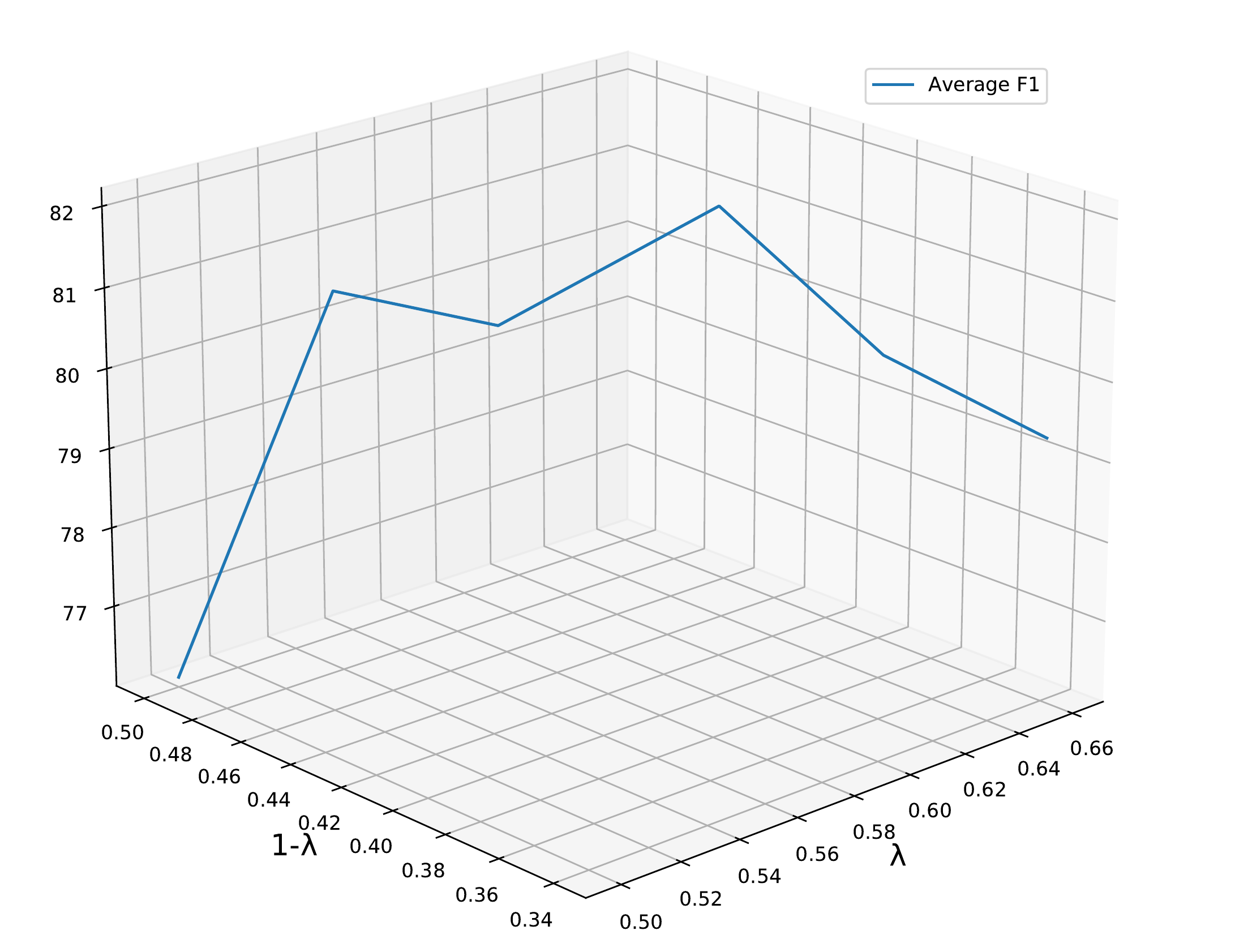}
\caption{ Sensitivity analysis of EEFRRD-SAEI model. The values of line is average $F1(\%)$ result from model with different $\lambda$.} \label{fig:Sen}
\end{figure}

We research the influence of the hyper-parameters  $\lambda$ with EEFRRD-SAEI model on the CFA dataset. We set $\lambda$ from 0.50 to 0.66 and test the performances of the proposed model under all the settings. The result is shown in Figure \ref{fig:Sen}, we observe that the performance of our model peaks around 0.6.

\subsection{Case Study}
\begin{figure}
\centering
\subfigure[Oracle]{
\begin{minipage}[t]{.5\textwidth}
\centering
\includegraphics[width=1\textwidth]{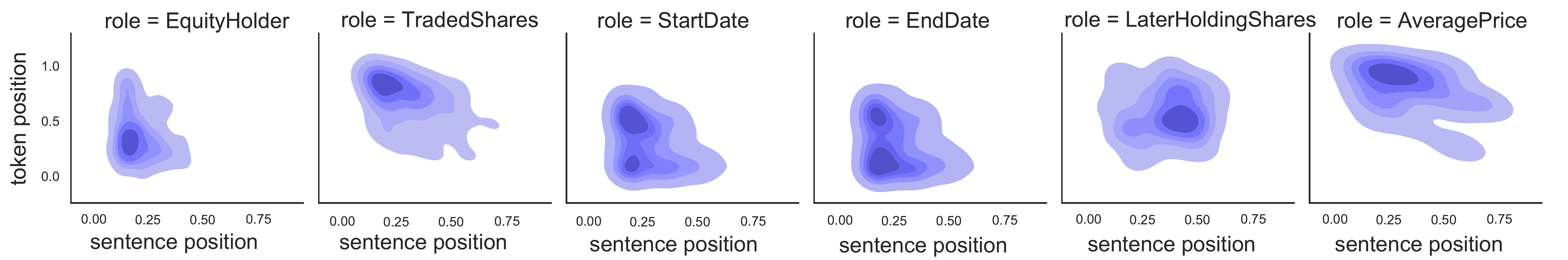}
\end{minipage}
}

\subfigure[DEE framework]{
\begin{minipage}[t]{.5\textwidth}
\centering
\includegraphics[width=1\textwidth]{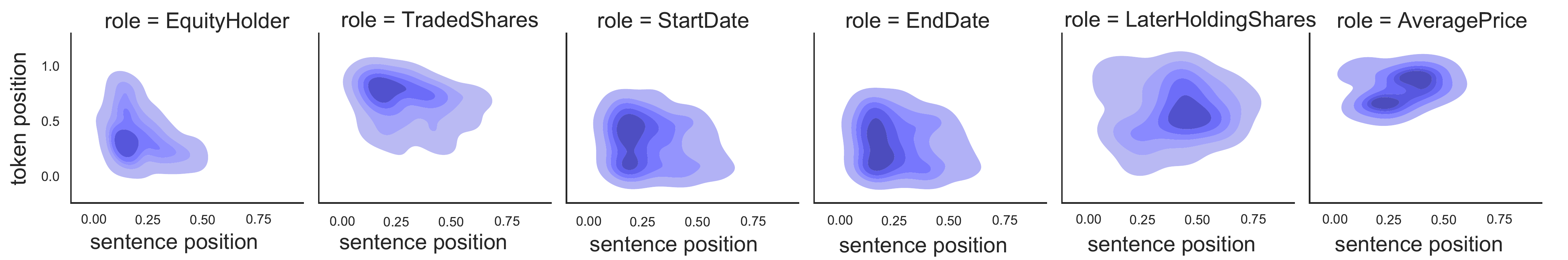}
\end{minipage}%
}%

\subfigure[Our model]{
\begin{minipage}[t]{.5\textwidth}
\centering
\includegraphics[width=1\textwidth]{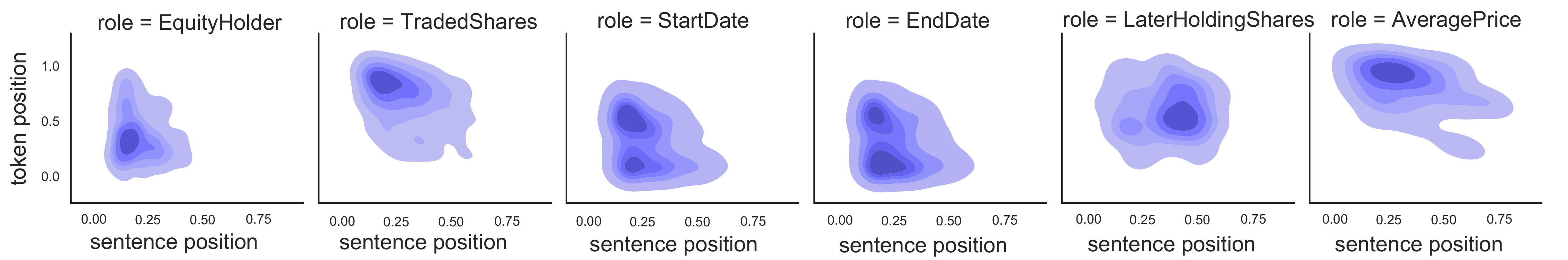}
\end{minipage}
}
\centering
\caption{An illustration of the Role-Rank Distribution (RRD) distributions in a financial news corpus (CFA, 2008-2018~\citep{10}). (a) shows the RRD distributions with the ground-truth annotations. (b) is plot with the result from a state-of-the-art model (DCFEE), and (c) is the result from our model. With the help of the proposed Role-Rank Distribution, the result of our method is more close to the ground-truth by comparing the result of the state-of-the-art method (b). }\label{fig4}
\end{figure}

To figure out the effectiveness of our model, we collect argument distribution from 64 EU documents as typical cases. (see Figure~\ref{fig4}) The "Oracle" is the corresponding ground-truth distribution in datasets. We observe that there are similar trends and shapes in both distribution curves figures of similar argument roles.

\section{Conclusion}
In this paper, we propose a document-level event extraction model to get the complete structural information from documents. Our method focuses on leveraging the writing style from the field-specific documents to improve the performance of the event extraction task. We formalize the writing style as the Role-Rank Distribution (RRD) that describes the relationship between the argument roles and the positions of the arguments in the sentences. Then, our model utilizes the proposed RRD in a self-attention module to revise the final event type prediction task. To further boost the efficiency of our model, we preprocess the RRD of each event type for the training data. The experimental results show that our model excels others on the performances of document-level event extraction tasks. This verifies that the writing style of documents helps to improve the performances to extract the event factors from the documents.

\bibliography{mybibfile}

\end{document}